\documentclass[a4paper,twoside,11pt]{article}

\usepackage[]{acl} 

\usepackage{times}
\usepackage{latexsym}

\usepackage[T1]{fontenc}

\usepackage[utf8]{inputenc}

\usepackage{microtype}

\usepackage{url}
\usepackage{graphicx} 
\usepackage{float} 
\usepackage{booktabs} 
\usepackage{amsmath}

\usepackage{array}
\usepackage{tabularx}
\usepackage{booktabs}
\usepackage[normalem]{ulem}
\useunder{\uline}{\ul}{}
\usepackage{booktabs} 

\usepackage{caption}
\usepackage{xcolor} 
\definecolor{mygreen}{RGB}{34, 139, 34}

\usepackage[utf8]{inputenc}
\usepackage{tcolorbox}
\tcbuselibrary{listings} 
\usepackage{geometry}
\usepackage{multicol} 
\usepackage{multirow}

\newtcolorbox[list inside=prompt,auto counter]{prompt}[1][]{
    colbacktitle=black!60,
    coltitle=white,
    fontupper=\footnotesize,
    boxsep=5pt,
    left=0pt,
    right=0pt,
    top=0pt,
    bottom=0pt,
    boxrule=1pt,
    #1,
}

\title{
Chain-of-Specificity: An Iteratively Refining Method for Eliciting Knowledge from Large Language Models
}

\author{Kaiwen Wei$^{1,2}$, Jingyuan Zhang$^{1}$, Hongzhi Zhang$^{1}$, Fuzheng Zhang$^{1}$, \\ \textbf{Di Zhang$^{1}$, Li Jin$^{2}$, Yue Yu$^{1}$}
\\
    $^{1}$Kuaishou Inc., China \\
    $^{2}$Key Laboratory of Network Information System Technology, Aerospace \\Information Research Institute, Chinese Academy of Sciences, China
\\ 
  \texttt{
  \{weikaiwen03,zhangjingyuan06\}@kuaishou.com} \\
}

\begin{document}
\maketitle


\begin{abstract}
    Large Language Models (LLMs) exhibit remarkable generative capabilities, 
    enabling the generation of valuable information.
    Despite these advancements, previous research found that LLMs sometimes struggle with adhering to specific constraints (e.g., \textit{in specific place} or \textit{at specific time}), at times even overlooking them, which leads to responses that are either too generic or not fully satisfactory.
    Existing approaches attempted to address this issue by decomposing or rewriting input instructions, yet they fall short in adequately emphasizing specific constraints and in unlocking the underlying knowledge (e.g., \textit{programming} within the context of \textit{software development}).
    In response, this paper proposes a simple yet effective method named Chain-of-Specificity (CoS).
    Specifically, CoS iteratively emphasizes the specific constraints in the input instructions, unlocks knowledge within LLMs, and refines responses. Experiments conducted on publicly available and self-build complex datasets demonstrate that CoS outperforms existing methods in enhancing generated content especially for the specificity. Besides, as the number of specific constraints increase, other baselines falter, while CoS still performs well.
    Moreover, we show that distilling responses generated by CoS effectively enhances the ability of smaller models to follow the constrained instructions.
    Resources of this paper will be released for further research.

\end{abstract}

\section{Introduction}


Large Language Models (LLMs) has marked a significant milestone in the field of natural language processing. The LLMs exhibit remarkable generative capabilities, 
which could generate meaningful information that was previously unattainable by earlier models \cite{DBLP:conf/naacl/DevlinCLT19, radford2019language}. They could be utilized in various fields, such as robotic systems \cite{DBLP:conf/aips/Kovalchuk0B21, DBLP:conf/icml/HuangAPM22} and problem-solving reasoning systems \cite{DBLP:conf/nips/Wei0SBIXCLZ22, DBLP:conf/iclr/0002WSLCNCZ23}.

\begin{figure}[t]
	\centering
	\includegraphics[width=7.8cm]{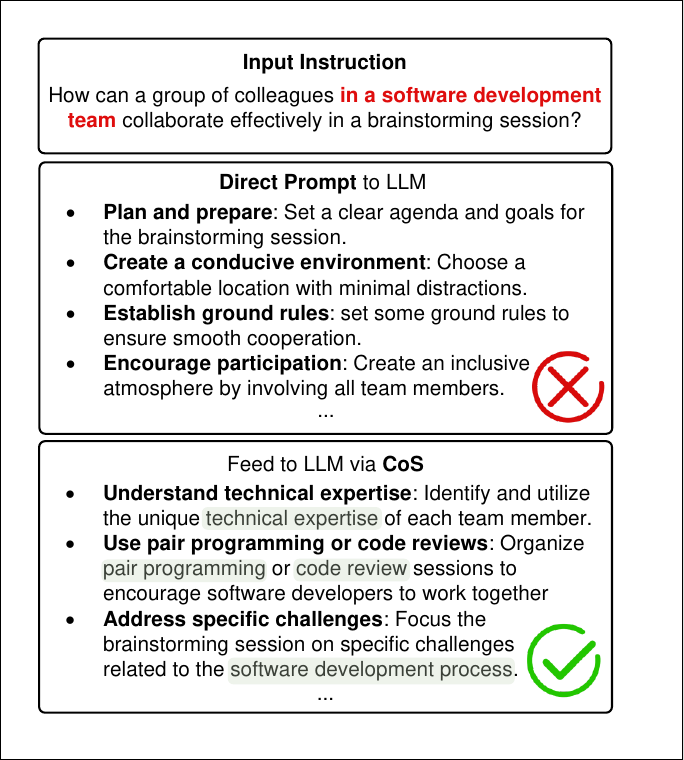}%
	\caption{The GPT-4 generation comparison between direct prompt method and Chain-of-Specificity (CoS). The direct prompt generate many generic responses, which could be broadly utilized in many other domain. In comparison, CoS generates more responses related to the specific constraint "\textit{software development team}". } 
	\label{fig:example}
\end{figure}

 Recent studies \cite{DBLP:conf/icml/HuangAPM22, DBLP:conf/emnlp/SakaguchiBBTCC21} primarily concentrate on devising plans for \textbf{general goals}, which akin to stereotypical activities described in \citet{abelson2014script}, such as "\textit{How can colleagues collaborate}". 
 Those methods have illustrated the proficiency of LLMs in generating a sequence of responses that align with the given instructions.
 However, \citet{DBLP:conf/acl/YuanCFGSJXY23}  found that LLMs sometimes fail to adhere strictly to \textbf{specific constraints}, which is defined as the multi-faceted and reasonable restrictions.
For example, as depicted in Fig.~\ref{fig:example}, even if we directly feed the prompt to the strong LLM GPT-4 \cite{openai2023GPT4}, it still struggle to grasp the specific constraint "\textit{software development team}". As a result, its responses are genetic and could be broadly utilized in many other domains, which dose not meet the requirement of the specific constraint.

However, how to address the issue of limited capacity in LLMs to capture specific constraints is under-exploit. 
There are methods such as decomposing input instructions into multiple sub-questions \cite{DBLP:conf/iclr/ZhouSHWS0SCBLC23, wang2023plan} and rewriting the input instructions to improve understanding \cite{DBLP:journals/corr/abs-2311-04155, DBLP:journals/corr/abs-2311-04205}, but these approaches exhibit limitations. Concretely, they fail to directly guide the model in comprehending the nuances of specific constraints. Furthermore, they overlook the exploration of the underlying knowledge within  these constraints. For instance, the domain of \textit{programming} is intricately linked to the context of \textit{software development}. \looseness=-1

Motivated by the findings in \citet{DBLP:conf/iclr/0002IWXJ000023} that LLMs contain enough knowledge for knowledge-intensive tasks, we introduce the Chain-of-Specificity (CoS) method to elicit the knowledge in LLMs and strengthen the ability of LLMs to follow the specific constraints. Specifically, 
it first identify the general goal and all the specific constraints in the input instruction.  After that, it takes the specific constraints as the reasoning chain and iteratively emphasises on the specific constraints to elicit the knowledge embedded in LLMs, and then revises the responses. As illustrated in Fig.~\ref{fig:example}, with the CoS method, the responses contains more information (e.g., \textit{code review}) about the specific constraint "\textit{software development team}".


In the experiment, we evaluate the methods on the CoScript~\cite{DBLP:conf/acl/YuanCFGSJXY23} dataset and the brainstorming domain of the EXPLORE-INSTRUCT dataset~\cite{DBLP:conf/emnlp/WanHYQB023} to validate the effectiveness of the proposed CoS method. Considering the limited quantity of specific constraints in those datasets,  we further developed a new dataset named ConstrainSPEC. Both machine evaluation and human assessment have corroborated that CoS achieves superior performance in specific constraint environments. 
Notably, CoS still perform well as the number of specific constraint increases.
In addition, we also conduct experiments on distilling the responses from different methods in ConstrainSPEC to smaller models, where the beat rate between those with CoS and those without distilling has reached 90.0.
In summary, the contributions of this paper are:

1) We propose the Chain-of-Specificity (CoS) method by iteratively eliciting the knowledge embedded in LLMs and 
refining the output responses for the specific constraints from the instructions. 

2) To stimulate the sophisticated  constraint situation, we develop a new dataset named ConstrainSPEC, which contains more and complex  specific constraints than other datasets.

3) We conduct experiments on the the relevant datasets. Both human and automatic evaluation illustrates the effectiveness of the CoS method. By leveraging the responses of different methods on LLMs, we endow the smaller models with better constrained instruction following ability.


\section{Related Work}
\subsection{LLMs under Constrained Situations}
Previous work \cite{DBLP:conf/icml/HuangAPM22} has shown that large language models (LLMs), such as GPT-3 \cite{DBLP:conf/nips/BrownMRSKDNSSAA20}, PaLM \cite{chowdhery2023palm}, and GPT-4 \cite{openai2023GPT4} can effectively generate the answers based on the input instructions in a zero/few-shot manner. 
Meanwhile, a wide range of works \cite{DBLP:conf/icml/HuangAPM22, DBLP:journals/corr/abs-2111-09276} focus on generating results for stereotypical activities toward general goals. 
There is only few work focus on discussing the ability of LLMs under constrained situations.
\citet{DBLP:conf/acl/YuanCFGSJXY23}  collected a dataset named coScript via overgenerate-then-filter, and then distill it to a smaller model. 
However, there are still some limitations: to begin with, the specific constraint number in   coScript is limited, which is not quite suit for simulate the complex constrained instruction situations. 
Additionally, they only evaluated on the scripts domain, while our work expand to brainstorming aspect with more specific constraints. This shift is driven by the intuition that 
brainstorming tasks involve more and broader knowledge, and are more difficult and realistic.

\begin{figure*}[!t] 
	\centering
	\includegraphics[width=15.5cm]{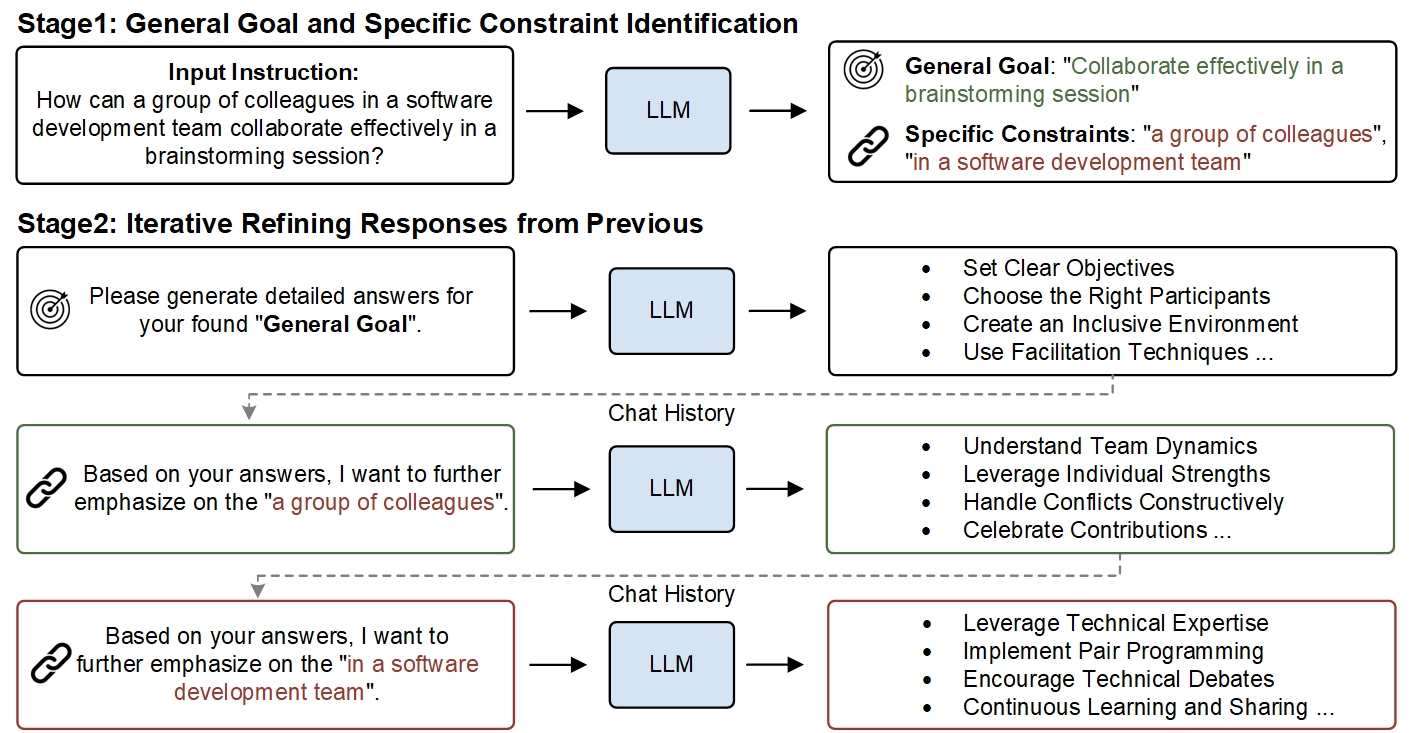}
	\caption{The overview of the proposed Chain-of-Specificity (CoS). }
	\label{model} 
\end{figure*}

\subsection{Methods under Constrained Situations}
\citet{DBLP:conf/acl/YuanCFGSJXY23} observed that LLMs sometimes do not adhere to the specific constraints. 
There are some methods \cite{DBLP:conf/iclr/ZhouSHWS0SCBLC23, DBLP:journals/corr/abs-2309-06275, wang2023plan} focus on 
breaking down the chain from the input instructions and then solving the sub-problems.
Besides, some methods \cite{DBLP:journals/corr/abs-2311-04155, DBLP:journals/corr/abs-2311-04205} seek to rewrite the input instructions to promote the understanding.
However, those methods did not explicitly direct the LLMs to follow the specific constraints in the input instructions, and unlock their  underlying knowledge. 
Based on the observation from \citet{DBLP:conf/iclr/0002IWXJ000023} that LLMs
contain enough knowledge for knowledge-intensive tasks, 
we proposed Chain-of-Specificity (CoS). It takes the specific constraints as the chain's backbone and elicits the knowledge embedded in LLMs by iteratively emphasising on the specific constraints.



\section{Method}
\subsection{Preliminary}
In this section, we will elucidate several pertinent terminologies. A  \textbf{general goal}, refers to stereotypical activities such as \textit{"How can colleagues collaborate"}.  A specific goal can be multi-facet with a reasonable constraint, such as "How can colleagues \textit{in a software development team} collaborate". Different from the name in the definition, we substituted 'specific goal' with '\textbf{specific constraint}' because in the experiment we found LLMs struggle to comprehend the words 'specific goals' in the input instructions. 

\subsection{Chain of Specificity (CoS)}
To tackle the challenge that LLMs sometimes neglect the specific constraints within input instructions and respond with general or even wrong results, we introduce a simple yet effective method named "Chain-of-Specificity" (CoS). 
As shown in Fig.~\ref{model}, the CoS encompasses two stages: 
(1) General goal and specific constraint identification, which aims at identifying the general goal and specific constraints within the input instruction, and 
(2) Iterative refining the responses from previous chat histories, which starts by generating a standard answer targeting the general goal, and then iteratively incorporates the underlying knowledge from specific constraints into the answers. 

\noindent\begin{minipage}{0.48 \textwidth} 
\begin{prompt}[title={General Goal and Specific Constraint Identification Prompt Template}]
You are asked to find the “General Goal” and the “Specific Constraints” based on the input Prompt. \\

\#\# Definition: \\
- A “General Goal” refers to stereotypical activities ... \\
- A “Specific Constraint” is derived from the corresponding general goal with various constraints ... \\
...
\\

\#\# Example: \\
Prompt: {Brainstorm 3 innovative advertising ideas for a new product launch targeting college students.} \\
- The “General Goal” is ... \\
- The “Specific Constraints” are ...\\
\\
\#\# Input Prompt: \\
\{\textless input\textgreater\} \\
\\
Answer in JSON. ...
\end{prompt}
\vspace{-10pt} 
\captionof{table}{The prompt template for identifying general goal and specific constraints, where \textless input\textgreater \  are the input prompt.}
\vspace{10pt} 
\label{stage1_prompt}
\end{minipage}

In the first stage, CoS scrutinizes the input instruction to discern the general goal and the specific constraints. Take the example in Fig.~\ref{model} as an example, given the input instruction, CoS initially identifies the general goal as \textit{"Collaborate effectively in a brainstorming session"}, while the specific constraints are recognized as \textit{"a group of colleagues"} and \textit{"in a software development team"}. The whole process is processed by asking the LLMs (e.g., GPT-4) and the example key structure  prompt template is shown in Table~\ref{stage1_prompt}. 


%

\noindent\begin{minipage}{0.48 \textwidth} 
\begin{prompt}[title={Prompt Template for General Goal Answers}]
Please generate detailed answers for your found "General Goal". The output should be as much elaborate as possible and in raw text format. Please provide a point by point description.
\end{prompt}
\vspace{-10pt} 
\captionof{table}{The prompt template for generating answers for general goal.}
\vspace{10pt} 
\label{stage2_general}
\end{minipage}

\noindent\begin{minipage}{0.48 \textwidth} 
\begin{prompt}[title={Prompt Template for Adding Specific Constraint}]
Based on your answers, I want to further emphasize on the "\textless Specific\_constrain\textgreater". Please regenerate the detailed answer based on the former answers in text format. Please provide a detailed point by point description and do not respond any other content.
\end{prompt}
\vspace{-10pt} 
\captionof{table}{The prompt template for appending the specific constraints to the answers, where \textless Specific\_constrain\textgreater \ are placeholder for the identified specific constraint in the first stage.}
\vspace{10pt} 
\label{stage2_specific}
\end{minipage}

\begin{table}[b]
\small
\centering
\begin{tabular}{ccc}
\toprule[1pt]
Dataset  & Methods          & General Scores \\ \hline
\multirow{2}{*}[-0.2ex]{CoScript}                 & Direct prompt    & 4.86           \\
                                                  & CoS-multi-step   & 4.84           \\  \hline
\multirow{2}{*}[-0.5ex]{\shortstack{EXPLORE \\ INSTRUCT}} & Direct prompt    & 4.68           \\
                                                  & CoS-multi-step   & 4.75           \\ 
\bottomrule[1pt]
\end{tabular}
\caption{The automatic evaluation results of general scores on two public datasets via GPT-4. }
\label{general_eval_EXPLORE_INSTRUCT}
\end{table}

In the second stage, the process begins with the LLMs generating a set of diverse, general answers that align with the identified general goal. This ensures a broad coverage of potential answers. The prompt template for generating answers for the identified general goal is shown in Table~\ref{stage2_general}. 
Subsequently, the method involves iteratively refining these answers by integrating one specific constraint at a time. The prompt template for incorporating various specific constraints could be found in Table~\ref{stage2_specific}.
Each round of CoS will further add emphasis on a specific constraint, while retaining the previous generation answers. This iteration will be stopped until all the specific constraints have been emphasised. 

In CoS, we could ask the LLMs to generate intermediate results at once through a single round of dialogue, or we can gradually let the LLMs emphasize specific constraint through multiple rounds of dialogue. Please find the whole prompt from CoS in Appendix~\ref{appendix:cos_one_step} and \ref{appendix:cos_multi_step}.
\looseness=-1



\begin{table}[t]
\small
\centering
\begin{tabular}{cc}
\toprule[1pt]
  Dataset               & \begin{tabular}[c]{@{}c@{}}Average Specific \\ Constraint Num\end{tabular}
   \\ \hline
coScript        & 1.00    \\ 
EXPLORE-INSTRUCT & 1.34                           \\
ConstrainSPEC       & 2.32                           \\ 
\bottomrule[1pt]
\end{tabular}
\caption{The specific constraint number comparison between different datasets, where ConstrainSPEC contains more specific constraints. }
\label{dataset_statistics}
\end{table}

\section{ConstrainSPEC with More Specific Constraints}

\subsection{Pilot Experiment}
To assess the models' comprehension of specific constraints, we initially select the coScript~\cite{DBLP:conf/acl/YuanCFGSJXY23} and the brainstorming domain in EXPLORE-INSTRUCT \cite{DBLP:conf/emnlp/WanHYQB023} as our evaluation datasets. The experimental results  presented in Table~\ref{general_eval_EXPLORE_INSTRUCT} reveal that inputting the raw prompt (direct prompt) into GPT-4 without any additional mechanisms, yields impressive results. This suggests that these two datasets are not particularly challenging, and GPT-4 is able to  accurately interpret the specific constraints in their instructions.
To delve deeper into the nature of these specific constraints, we quantify the average number of specific constraints present in both datasets. 
Specifically, we employ the prompt template shown in Table~\ref{stage1_prompt} to determine the number of specific constraints in each instruction, eventually calculating the average per instruction.
The experiment is shown in Table~\ref{dataset_statistics}, which indicates that both coScript and EXPLORE-INSTRUCT contain averagely only about one specific constraint. 
All those findings demonstrate existing datasets lack of a substantial number of specific constraints, rendering them inadequate for simulating scenarios with complex and multiple specific constraints.


\subsection{Dataset Construction}
To address the limitations identified earlier and more rigorously test the methods in intricate scenarios, particularly those involving numerous specific constraints, we develop a new dataset named ConstrainSPEC. 
It is constructed as follows: we first randomly select 1,000 instructions from the brainstorming section of the EXPLORE-INSTRUCT dataset, and then we ask LLMs to enhance these instructions, infusing them with a greater complexity and a higher number of specific constraints.
The example template used for dataset construction is outlined in Table~\ref{dataset_construct}. See Appendix~\ref{appendix:Dataset_consturct} for the detailed prompt. 
We regard those generated 1,000 samples as the test set of ConstrainSPEC.

\noindent\begin{minipage}{0.48 \textwidth} 
\begin{prompt}[title={Prompt Template for Dataset Construction}]
You are asked to add certain reasonable constraints to the input prompt. The modified prompt requires the models to pay attention to relevant details after retrieving certain background knowledge. 
\\

\#\# Guidelines \\
- You should create an appropriate and logical modified prompt based on the input prompt. \\
- The response you generated should conform in json format.
\\

\#\# Examples: \\
\textless Example1\textgreater \\
- Input: Render a 3D model of a house. \\
- Modified: Render a 3D model of a house for a senior citizen. \\
- Reason: I append a constraint “for a senior citizen”. The reasons are as follows: because when designing a house, compared with normal young people, the elderly need extra care, such as designing electric stairs. \\
... \\

\#\# Input prompt \\
\{\textless input\_sentence\textgreater\} \\

List one modified prompt examples of the above input prompt. 
\end{prompt}
\vspace{-10pt} 
\captionof{table}{Dataset construction  template, where \textless input\_sentence\textgreater \ means the raw input instruction.}
\vspace{10pt} 
\label{dataset_construct}
\end{minipage}

\vspace{-5pt}
\subsection{Dataset Analyse}
\begin{figure}[t!] 
	\centering
	\includegraphics[width=7.8cm]{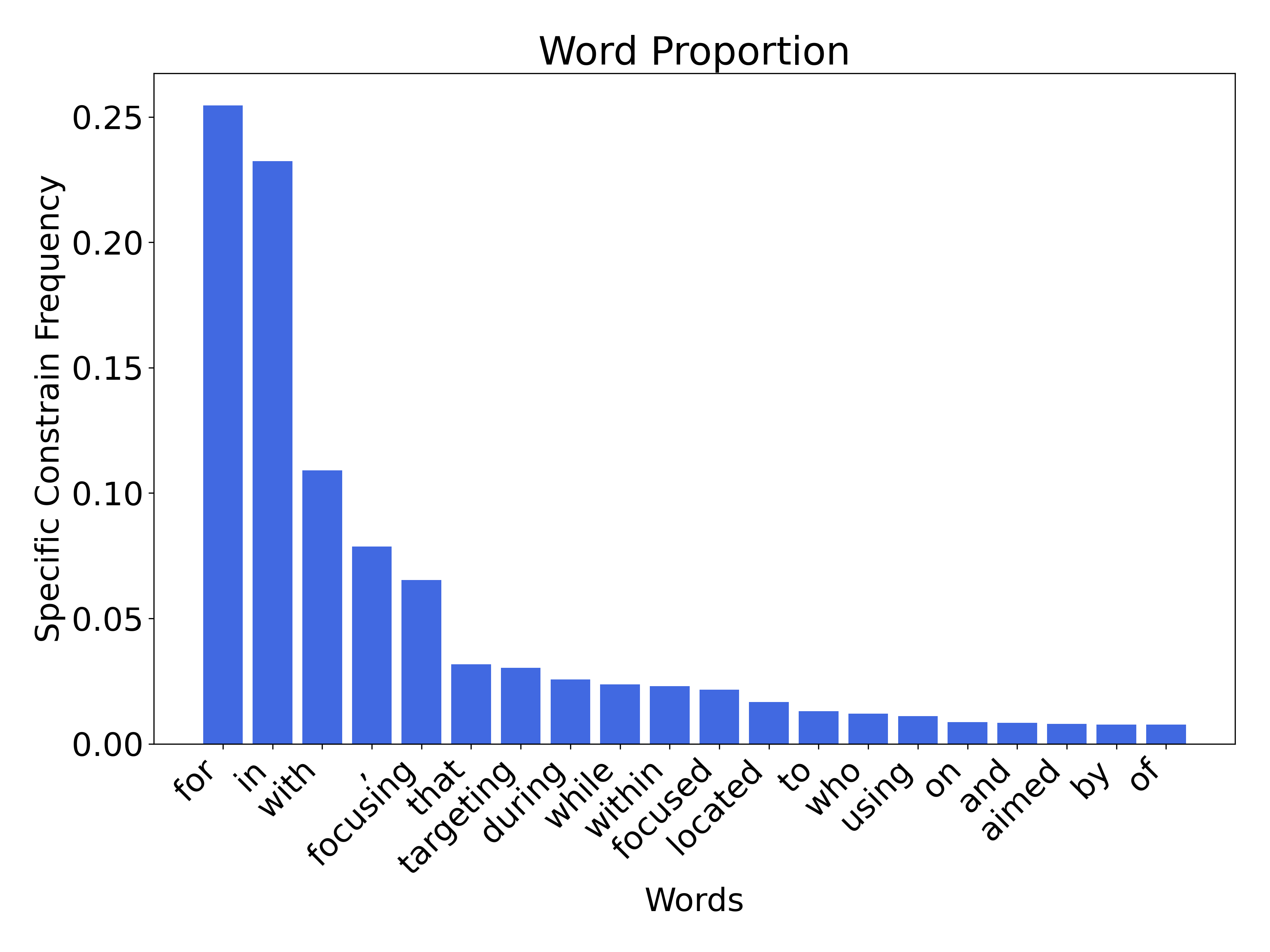}
	\caption{The initial words of the added  specific constraints in ConstrainSPEC test set. }
	\label{top20_specific} 
\end{figure}

As shown in Table~\ref{dataset_statistics}, the averaged specific constraint number of ConstrainSPEC is higher than the other two datasets. 
To better showcase its statistics, we conduct a detailed analysis on the added specific constraints. Specifically, following \citet{DBLP:conf/acl/YuanCFGSJXY23}, we visualized the data by plotting the initial word of the top 20 added  specific constraints. As shown in Fig.~\ref{top20_specific}, we could find a significant portion of the added specific constraints pertains to \textit{intent} (e.g., \textit{for}) or \textit{method} (e.g., \textit{in} or \textit{with}) categories according to the taxonomy in Probase \cite{DBLP:conf/sigmod/WuLWZ12}.
Moreover, there is a notable prevalence of subordinate clauses, as indicated by the frequent use of commas, the word "that", and other similar linguistic markers. This suggests that constraints are semantically specific and syntactically complex.


\vspace{-3pt}
\section{Distilling to Smaller Models}
As demonstrated in Fig.\ref{fig:example}, the advanced GPT-4 model still faces challenges in adhering to specific constraints, a problem that is accentuated in smaller-scale models. This issue is further evidenced by the experiments shown in Table~\ref{general_eval_ConstrainSPEC} and Table~\ref{human_general_eval_ConstrainSPEC}, which reveals the struggles of smaller LLMs like Vicuna-13b \cite{DBLP:journals/corr/abs-2306-05685} and Llama2-Chat-13b \cite{DBLP:journals/corr/abs-2307-09288} in grasping specific  constraints. In this section, we aim to augment these smaller LLMs' capabilities to respect such constraints more effectively.

To this end, we generate 5,000 samples using the dataset construction template outlined in Table~\ref{dataset_construct} on the EXPLORE-INSTRUCT dataset, and set them as  the training set of ConstrainSPEC. Please note that there is no overlap between these 5,000 samples and the generated test set. We then feed the ConstrainSPEC training dataset to larger LLMs (e.g., GPT-4) and let them generate responses through two prompt  methods: (1) \textbf{CoS-multi-step prompt}, employing the proposed CoS method with multiple reasoning steps; (2) \textbf{direct prompt}, directly inputting the instructions. After that, the responses of larger LLMs  generated from these methods are subsequently used for training smaller LLMs via supervised fine-tuning. 
\vspace{-7pt}
\section{Experiment}



\vspace{-5pt}
\subsection{Baseline}
In the experiment, we leverage GPT-4 \cite{openai2023GPT4} with the \textit{gpt-4-1106-preview} version as the base LLM and we compare with the strong baselines:
(1) \textbf{Direct prompt}: Naive prompting to generate the responses; 
(2) \textbf{CoT} \cite{DBLP:conf/nips/Wei0SBIXCLZ22}: Automatic generation of series of intermediate reasoning steps from LLMs with prompt "\textit{let's think step by step}"; 
(3) \textbf{Take-a-breath} \cite{DBLP:journals/corr/abs-2309-03409}: Enhanced CoT by prompting "\textit{Take a deep breath}";
(4) \textbf{Least-to-Most} \cite{DBLP:conf/iclr/ZhouSHWS0SCBLC23}: First automatically decomposing the inhand problems into series of simpler sub-problems, and then each one  sequentially;
(5) \textbf{Plan-and-Solve} \cite{wang2023plan}: Enhanced CoT by guiding LLMs to devise the plan before solving the problems;
(6) \textbf{Re-Reading} \cite{DBLP:journals/corr/abs-2309-06275}: Entails revisiting the question information embedded within input prompts;
(7) \textbf{RaR-one-step} \cite{DBLP:journals/corr/abs-2311-04205}: Rephrase and expand questions posed by humans and provide responses in a single prompt in a single response;
(8) \textbf{RaR-multi-step} \cite{DBLP:journals/corr/abs-2311-04205}: Rephrase the question and respond the rephrased question in multiple steps;
(9) \textbf{BPO} \cite{DBLP:journals/corr/abs-2311-04155}:  Rewrite user prompts to suit LLMs’ input understanding;
(10) \textbf{CoS-one-step}: The proposed Chain-of-specificity (CoS) method that combines identifying general goal, specific constraints, and adding the specific constraints to the answers in a single response;
(11) \textbf{CoS-multi-step}: The proposed CoS method iteratively adds the specific constraints to the answers in different steps at different stages.


\begin{figure*}[!t] 
	\centering
	\includegraphics[width=15.5cm]{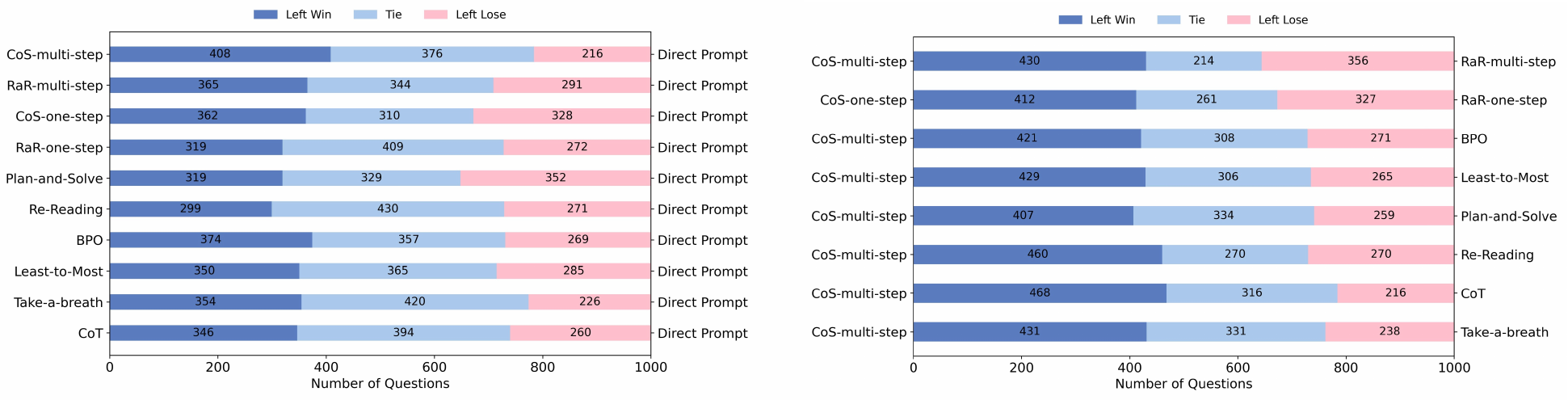}
	\caption{The pairwise automatic evaluation results on ConstrainSPEC test set. }
	\label{auto_eval} 
\end{figure*}
\begin{table}[]
\small
\centering
\begin{tabular}{ccc}
\toprule[1pt]
Methods                                                                      & Win:Tie:Lose & Beat Rate \\ \hline
\begin{tabular}[c]{@{}c@{}}CoS-one-step vs \\ Direct prompt\end{tabular}   & 287:567:146      & 66.3          \\ \hline
\begin{tabular}[c]{@{}c@{}}CoS-multi-step vs \\ Direct prompt\end{tabular} & 333:524:143      & 69.5          \\ 
\bottomrule[1pt]
\end{tabular}
\caption{The pairwise automatic evaluation results on the EXPLORE-INSTRUCT dataset. }
\label{beat_rate_EXPLORE_INSTRUCT}
\end{table}
\begin{table}[]
\small
\centering
\begin{tabular}{cc}
\toprule[1pt]
Methods                             & General Scores \\ \hline
\multicolumn{2}{c}{\textit{Vicuna-13b}}                                              \\ \hline
Direct prompt                       & 3.82 
\\ \hline
\multicolumn{2}{c}{\textit{Llama2-Chat-13b}}                                              \\ \hline
Direct prompt                       & 4.23 
\\ \hline
\multicolumn{2}{c}{\textit{GPT-4}}                                              \\ \hline
Direct prompt                       & 4.47       \\
CoT                                 & 4.54       \\
Take-a-breath                       & 4.55       \\
Re-Reading                          & 4.51       \\
Plan-and-Solve                      & 4.59       \\
Least-to-Most                     & 4.57       \\
BPO                & 4.63       \\
RaR-one-step       & 4.52       \\
CoS-one-step       & 4.59       \\
RaR-multi-step      & 4.66       \\
CoS-multi-step     & 4.80       \\ 
\bottomrule[1pt]
\end{tabular}
\caption{The automatic evaluation results of general scores on the ConstrainSPEC dataset. }
\label{general_eval_ConstrainSPEC}
\end{table}

\vspace{-5pt}
\subsection{Automatic Evaluation}
To evaluate the performance of the methods, we follow \citet{DBLP:journals/corr/abs-2304-10453} and \citet{DBLP:conf/emnlp/WanHYQB023} to conduct an automatic evaluation with GPT-4. Specifically, we adopt (1) \textbf{general  scores evaluation} (1 for the worst and 5 for the best), which aims to capture the qualities of the generated results. Please refer to Appendix~\ref{appendix:Overall_Scores_Evaluation} for the prompts and the standards used to solicit scores.
(2) \textbf{pair-wise evaluation}, where given an instruction and two responses from different methods, we request GPT-4 to determine which response is better based on their understanding of the general goal and specific constraints. Refer to Appendix~\ref{appendix:Pairwise_Scores_Evaluation} for the prompt for pairwise evaluation. 
Moreover, to calculate the beat rate of a particular model, we divide the number of times the model wins by the sum of the number of times the model wins and loses. 
Please refer to Appendix~\ref{auto_evaluation_details} for more details about the automatic evaluation settings. \\
\textbf{Experiments on EXPLORE-INSTRUCT.}
We conduct the experiments on the EXPLORE-INSTRUCT to exam the generalization of CoS. The experiment results are shown in Table~\ref{general_eval_EXPLORE_INSTRUCT} and Table~\ref{beat_rate_EXPLORE_INSTRUCT}. We could find that both the direct prompt and CoS method show great performance on the EXPLORE-INSTRUCT dataset. Meanwhile, compared to the experiment results on ConstrainSPEC in Table~\ref{general_eval_ConstrainSPEC} that containing more specific constraints, the general score of direct prompt on EXPLORE-INSTRUCT is much higher. 
Those findings supports our hypothesis that the original datasets, with its limited number of specific constraints, may not adequately simulate more complex scenarios. Furthermore, the experimental results also indicate that the CoS method outperforms the direct prompt approach. This underscores CoS's robustness and adaptability in scenarios where the number of specific constraints is inherently limited. 
\begin{table}[t!]
\small
\centering
\begin{tabular}{ccc}
\toprule[1pt]
Methods                                                                      & Win:Tie:Lose     & Beat Rate \\ \hline
\multicolumn{3}{c}{\textit{Vicuna-13b}}                                              \\ \hline 
\begin{tabular}[c]{@{}c@{}}CoS-multi-step vs \\ Direct prompt\end{tabular}     & 402:280:318      & 55.8      \\ \hline
\begin{tabular}[c]{@{}c@{}}CoS-multi-step vs \\ w/o distill\end{tabular}     & 659:268:73       & 90.0      \\ \hline
\begin{tabular}[c]{@{}c@{}}Direct prompt vs \\ w/o distill\end{tabular}  & 668:205:127      & 84.0      \\ \hline
\multicolumn{3}{c}{\textit{Llama2-Chat-13b}}                                              \\ \hline 
\begin{tabular}[c]{@{}c@{}}CoS-multi-step vs \\ Direct prompt\end{tabular}     & 373:310:317      & 54.0      \\ \hline
\begin{tabular}[c]{@{}c@{}}CoS-multi-step vs \\ w/o distill\end{tabular}     & 437:332:231       & 65.4      \\ \hline
\begin{tabular}[c]{@{}c@{}}Direct prompt vs \\ w/o distill\end{tabular}  & 405:331:264      & 60.5      \\ 
\bottomrule[1pt]
\end{tabular}
\caption{The pairwise automatic evaluation results on distilling for two smaller LLMs. }
\label{distill_result}
\end{table}
\begin{figure}[t!] 
	\centering
	\includegraphics[width=6.0cm]{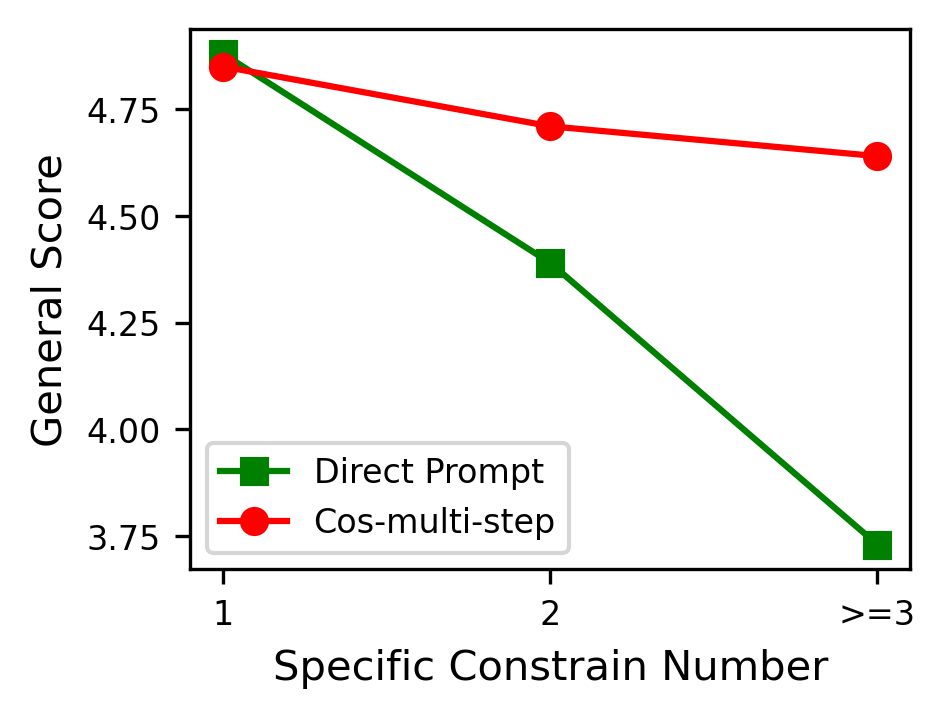}
	\caption{The automatic evaluated general scores with different specific constraint numbers. }
	\label{specific_cons_num} 
\end{figure}
\begin{figure*}[!t] 
	\centering
	\includegraphics[width=15.5cm]{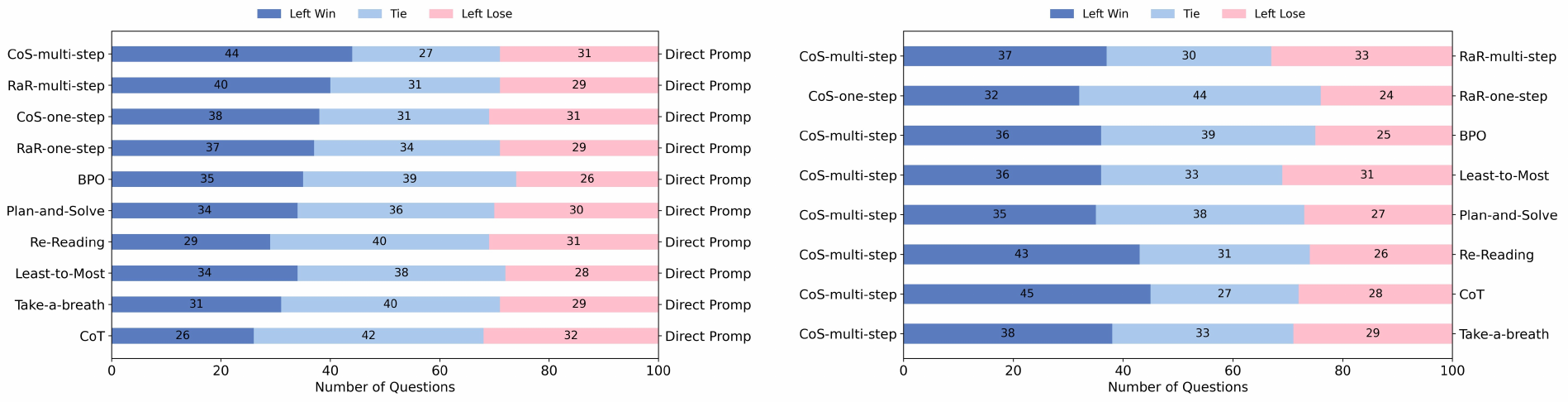}
	\caption{The pairwise human evaluation results on ConstrainSPEC test set. }
	\label{human_eval} 
\end{figure*}
\\
\textbf{Experiments on ConstrainSPEC.}
We conduct experiments on the test set of ConstrainSPEC dataset, which is more complex and has more specific constraints. From the experiment results in Table~\ref{general_eval_ConstrainSPEC}, we could observe that (1) CoS outperforms other strong methods, indicating its superiority in complex specific constraint situations; (2) The promotion of those methods such as CoT is not significant. This is possibly because that it tent to generate 
intermediate results while skimming over specific responses; (3) Those methods utilizing the multi-step for generating answering typically have greater general scores. A key reason is that they could consider the history information during generation. 
In addition, the results in Fig.~\ref{auto_eval} also reveal that every baseline outperforms the direct prompt, and the proposed CoS method has greater beat rate other strong methods. For example, the beat rate of CoS-multi-step vs direct prompt is 65.4\%, showing the superiority of CoS in the situation with complex specific constraints.
\\
\textbf{Experiments with Different Specific Constraint Number.}  
As shown in Fig.~\ref{specific_cons_num}, we explored the model's performance across various numbers of specific constraints on the ConstrainSPEC test set. It can be observed that while the direct prompt approach achieved commendable performance when the number of specific constraints was limited to one, its performance significantly deteriorated with the increase in the number of specific constraints. However, the CoS-multi-step approach maintained a relatively stable performance across different specific constraint settings, demonstrating the effectiveness of the CoS method under complex specific constraint situations.
\\
\textbf{Experiments on Distilling to Smaller Models.}
We conduct experiments on distilling  knowledge from larger LLMs to the smaller LLMs. We select Vicuna-13b \cite{DBLP:journals/corr/abs-2306-05685} and Llama2-Chat-13b \cite{DBLP:journals/corr/abs-2307-09288} since they are typical smaller LLMs. 
We employ two prompt strategies on GPT-4 to generate the responses: (1) {CoS-multi-step}, (2) {direct prompt}, and we also provide (3) {w/o distill}, where the smaller LLMs are tested directly without distillation. 
The detailed distillation experiment settings are shown in Appendix~\ref{distill_setting}.
The results in Table~\ref{distill_result} on the ConstrainSPEC test set indicate: compared to w/o disll, other methods have marked promotion in the smaller models' capabilities to adhere to constrained instructions, validating the effectiveness of the distillation strategy and the responses quality from different prompt methods. Moreover, the data shows a beat rate of 55.8\% favoring the CoS-multi-step over direct prompting, signifying the superiority of the CoS methods' responses in guiding smaller models toward more accurate compliance with specific instructions. 

\begin{table}[]
        \small
\centering
\begin{tabular}{cc}
\toprule[1pt]
Methods                             & General Scores \\ \hline
\multicolumn{2}{c}{\textit{Vicuna-13b}}                                              \\ \hline
Direct prompt                       & 3.51 
\\ \hline
\multicolumn{2}{c}{\textit{Llama2-Chat-13b}}                                              \\ \hline
Direct prompt                       & 4.09 
\\ \hline
\multicolumn{2}{c}{\textit{GPT-4}}                                              \\ \hline
Direct prompt                       & 4.34       \\
CoT                                 & 4.26       \\
Take-a-breath                       & 4.38       \\
Re-Reading                          & 4.37       \\
Plan-and-Solve                      & 4.36       \\
Least-to-Most                       & 4.50       \\
BPO                & 4.55     \\
RaR-one-step            & 4.52       \\
CoS-one-step       & 4.57       \\
RaR-multi-step & 4.59       \\
CoS-multi-step     & 4.69       \\ 
\bottomrule[1pt]
\end{tabular}
\caption{The human evaluation results of general scores on the ConstrainSPEC dataset. }
\label{human_general_eval_ConstrainSPEC}
\end{table} 

\vspace{-9pt}
\subsection{Human Evaluation}

For a thorough and unbiased evaluation, we randomly selected 100 ConstrainSPEC samples for human evaluation.
Specifically, we enlist three annotators to (1) Give a general score for each responses based on the same standard as automatic evaluation; (2) Compare responses from two methods to the same instruction, judging which model performed better (win, tie, or lose). To avoid bias, model identities were hidden, we also keep annotators blind to the source of each response.

Tabel~\ref{human_general_eval_ConstrainSPEC} and Fig.~\ref{human_eval} show the results of the human evaluation. 
The comparison between the human and automatic evaluations demonstrates a general consistency, where the Fleiss’s K \cite{fleiss1981measurement} is 0.73. It indicates that the automatic evaluation is also qualitatively well-regarded by humans, illustrating the reliability of the automatic evaluation. 
From the results, 
we could find that CoS demonstrates  superiority over other baselines, and notably, direct prompts exceed CoT in the brainstorming task.
After manually check the generated results in CoT, we find GPT-4 tend to break each point in the responses into many steps, which dose not contributes to increase the comprehension of input instructions. 
\\
\textbf{Case Study.}  
As shown in Table~\ref{case_study}, we select several typical cases to illustrate the effectiveness of the proposed CoS method. For example, when the input instruction contains specific constraint "\textit{in a healthcare setting}", the proposed CoS-multi-step successfully elicits the background knowledge in LLMs, and the contents such as "\textit{staff-to-patient ratios}" in the response are more in line with the specific constraint "\textit{healthcare}". Please refer to Appendix~\ref{appendix:case_study} for the full results.

\begin{table}[t!]
\centering
\footnotesize
\begin{tabular}{m{2cm}|m{5cm}}
\toprule[1pt]
\centering \textbf{Input Instruction} & \centering \textbf{CoS-multi-step} \tabularnewline
\toprule[1pt]
\centering Brainstorm ways to improve employee morale \textcolor{red}{in a healthcare setting}. &  1. Address staffing issues: Ensure appropriate \textcolor{mygreen}{staff-to-patient ratios} to reduce stress and burnout. ... 3. Cross-training opportunities: Encourage and offer \textcolor{mygreen}{cross-training opportunities for healthcare employees} to learn new skills, etc. ...\tabularnewline
\bottomrule[1pt]
\end{tabular}
\caption{Case study experiment. The specific constraints are in \textcolor{red}{red}, the relevant responses are in \textcolor{mygreen}{green}.}
\label{case_study}
\end{table}

\vspace{-8pt}
\section{Conclusion}
\vspace{-5pt}
To increase LLMs' ability to follow the specific constraints in the input instructions, we propose Chain-of-Specificity (CoS) 
by iteratively emphasising on the specific constraints, eliciting  knowledge in LLMs, and refining the responses. 
To better stimulate the complex constraint situations, we further propose a new dataset named ConstrainSPEC, containing more and complex specific constraints. 
Experiments on the public and self-build  datasets illustrate the effectiveness of CoS to direct LLMs to adhere to specific constraints. 
Moreover, the smaller models are equipped with better constrained instruction following ability by distilling the responses from CoS. 

\section{Limitation}
This paper expands the research from scripts domain to brainstorming domain. There are still numerous other areas still significantly require LLMs to adhere to specific constraints, including but not limited to story writing domain. 
In addition, the polot experiment illustrate that existing datasets lacks of a substantial number of specific constraints.
Due to financial limitations, we have collected 6,000 samples for the ConstrainSPEC dataset. This new dataset is tailored specifically to the brainstorming domain and introduces more and complex  specific constraints. We believe that the samples in ConstrainSPEC are sufficient for evaluating the models under the complex specific constraint scenarios. We leave the methods to alleviate those limitations as the future work.


\section{Ethics Statement}
Understanding the paramount importance of information security in the development and application of LLMs, our study prioritizes the ethical sourcing and handling of data. The source data for our research is derived exclusively from the open-source dataset EXPLORE-INSTRUCT, which is publicly available and does not contain any personally identifiable information or sensitive data. This approach ensures that our research adheres to privacy and data protection standards, minimizing risks associated with data misuse.

The potential for LLMs to generate content that could be considered toxic or harmful has been documented in previous studies. Acknowledging this risk, we have taken proactive measures to mitigate the possibility of such outcomes in our work. It is important to clarify that our dataset, while comprehensive, is not designed for use in safety-critical applications or as a substitute for specialized, expert advice in sensitive domains. The purpose of our dataset is to facilitate research and development in specific, non-critical areas of natural language processing.

To further ensure the integrity and safety of the data used in our study, annotators were given explicit instructions to identify and exclude any content that could be deemed offensive, harmful, or otherwise inappropriate during the review process of the test set in ConstrainSPEC. This careful curation process is part of our commitment to responsible research practices and contributes to the overall quality and reliability of our dataset.

Moreover, we explicitly state that any research outcomes or applications derived from this study are intended strictly for academic and research purposes. We do not authorize the use of our findings or the ConstrainSPEC dataset for commercial purposes without proper oversight and ethical considerations. 




\bibliography{anthology}

\newpage
\appendix
\newpage

\section{Appendix}
\subsection{Prompt Template for Chain of Specificity One Step}
\label{appendix:cos_one_step}

\noindent\begin{minipage}{0.48 \textwidth} 
\begin{prompt}[title={Prompt Template for Chain of Specificity One Step}]

\#\# Definition: \\
- A “General Goal” refers to stereotypical activities, e.g., “make a cake”. \\
- A “Specific Goal” is derived from the corresponding general goal with various constraints, e.g., “make a chocolate cake”. \\

\#\# Example: \\
- Input Prompt: \{Brainstorm 3 innovative advertising ideas for a new product launch targeting college students.\} \\
- The “General Goal” is to "Brainstorm ideas for a product launch", the “Specific Goal” are "3 innovative advertising ideas", "new product launch", and "targeting college students". \\

\#\# Note: \\
- The "General Goal" and "Specific Goal" MUST be found from the raw prompt. \\
- The "General Goal" is a short sentence that highly covers the main information required for input prompts. \\
- The "Specific Goal" needs to be as detailed as possible, and it must be found from the input prompt text. \\

\#\# Input Prompt: \\
\{<input>\} \\

\#\# Task: \\
- You will first generate as many answers as possible based on the “General Goal” of the above Prompt. Then generate specific, compatible with “Specific Goal” answers based the above Prompt. \\
- Repeat the following 2 steps several times. \\
  Step 1. Identify 1 Specific Goal from the Prompt which is semantically missing from the previously generated answer. \\
  Step 2. Write a new answer which covers the new identified Specific Goal. \\
  If you can't find any other Specific Goal, stop this iteration. \\
- Based on all the previously contents generated for the General Goal and Specific Goals, you need to generate the answer from the Input Prompt item by item. Expand each item and introduce it in detail. \\

\#\# Guidelines: \\
- The first answer should semantically cover the General Goal yet be highly non-specific. Generate as many answers as possible by sub-pointing. \\
- Give specific and compatible answer that is suitable for each Specific Goals. \\
- The written answer should be well-structured, with a logical flow of ideas and clearly defined sections or headings for different components of the answer. 
\end{prompt}
\vspace{-10pt} 
\vspace{10pt} 
\end{minipage}

\noindent\begin{minipage}{0.48 \textwidth} 
\begin{prompt}[title={Prompt Template for Chain of Specificity One Step}]

\#\# Output Format \\
Answer in JSON. The format is as follow: \\
\{ \\
"General Goal": \_\_\_\_\_, \\
"Specific Goal1": \_\_\_\_\_, \\
"Specific Goal2": \_\_\_\_\_, \\
... \\
"Answer": \_\_\_\_\_, \\
\} \\
\\
Contents in the \_\_\_\_\_ are all the raw text rather than other formats. The value for the first key "General Goal" is the answer for the general goal, and the values in the middle are answers for different "Specific Goals", the final value for the key "Answer" is the answer that generates results based on all previously contents generated for the General Goal and Specific Goals. In the last element, "Answer" should be raw text separated by numbers, each separated by a newline.

\end{prompt}
\vspace{-10pt} 
\captionof{table}{The prompt template for chain-of-specificity-one-step, where \textless input\textgreater \ are placeholder for the raw input prompt.}
\vspace{10pt} 
\end{minipage}

\subsection{Prompt Template for Chain of Specificity multi Step}
\label{appendix:cos_multi_step}
\noindent\begin{minipage}{0.48 \textwidth} 
\begin{prompt}[title={General Goal and Specific Constraints}]
\textbf{STEP1: } \\

\#\# Definition: \\
- The "General Goal" and "Specific Constraint" MUST come from the Prompt content. \\
- A “General Goal” refers to stereotypical activities, e.g., “make a cake”. It is highly non-specific and does not include any details. \\
- A “Specific Constraint” Is derived from the corresponding general goal with various constraints, e.g., “make a chocolate cake”. \\
- Please find the specific constraints as detail as possible. \\

\#\# Example: \\
Prompt: \{Brainstorm 3 innovative advertising ideas for a new product launch targeting college students.\} \\
- The “General Goal” is "Brainstorm ideas for a product launch". \\
- The “Specific Constraints” are "3 innovative advertising ideas", "new product launch", and "targeting college students". \\

\#\# Input Prompt: \\
\{<input>\} \\

Answer in JSON. The keys of the json are "General Goal" and "Specific Constraints". The value of "Specific Constraints" is a list that includes all the "Specific Constraints" that you find from the Prompt content. Make sure the "General Goal" and "Specific Constraints" are from the Prompt content.

\end{prompt}
\vspace{-10pt} 
\vspace{10pt} 
\end{minipage}

\noindent\begin{minipage}{0.48 \textwidth} 
\begin{prompt}[title={General Goal and Specific Constraints}]
\textbf{STEP2: } \\

Please generate detailed answers for your found "General Goal". The output should be as much elaborate as possible and in raw text format. Please provide a point by point description.
\\

\textbf{STEP3: } \\

Based on your answers, I want to further emphasize on the "\textless Specific\_constrain\textgreater". Please regenerate the detailed answer based on the former answers in text format. Please provide a detailed point by point description and do not respond any other content.

\end{prompt}
\vspace{-10pt} 
\captionof{table}{The prompt template for chain-of-specificity-multi-step, where \textless input\textgreater \ are placeholder for the raw input prompt and \textless Specific\_constrain\textgreater are the specific constraints that detected in STEP1.}
\vspace{10pt} 
\end{minipage}

\subsection{Prompt Template for Dataset Construction}
\label{appendix:Dataset_consturct}
\noindent\begin{minipage}{0.48 \textwidth} 
\begin{prompt}[title={Prompt Modification and Reasoning}]

\#\# Guidelines \\
- You should create an appropriate and logical modified prompt based on the input prompt. \\
- The response you generated should conform to the following json format: \\
\{ \\
  "Output1": \{ \\
    "Input": \_\_\_\_, \\
    "Modified": \_\_\_\_, \\
    "Reason": \_\_\_\_, \\
  \}, \\
  "Output2": \{ \\
    "Input": \_\_\_\_, \\
    "Modified": \_\_\_\_, \\
    "Reason": \_\_\_\_\_, \\
  \} \\
	... \\
\} \\
where "Input" is the input prompt, "Modified" is the prompt after modification, and the "Reason" is the detailed reason for why appending the constraints and what background knowledge is behind the constraints. \\

\#\# Examples: \\
\textless Example1\textgreater \\
- Input: Render a 3D model of a house. \\
- Modified: Render a 3D model of a house for a senior citizen. \\
- Reason: I append a constraint “for a senior citizen”. The reasons are as follows: because when designing a house, compared with normal young people, the elderly need extra care, such as designing electric stairs. \\

\end{prompt}
\vspace{-10pt} 
\vspace{10pt} 
\end{minipage}

\noindent\begin{minipage}{0.48 \textwidth} 
\begin{prompt}[title={Prompt Modification and Reasoning}]

\textless Example2\textgreater \\
- Input: Come up with possible solutions for improving office productivity. \\
- Modified: Come up with possible solutions for improving office productivity for a small startup. \\
- Reason: I append a constraint “for a small startup”. The reasons are as follows: because the small startup doesn't have sufficient financial strength, so compared to large companies, more cost-effective methods are needed to improve office productivity. \\
\textless Example3\textgreater \\
- Input: Identify methods to decrease absenteeism and improve employee engagement. \\
- Modified: Identify methods to decrease absenteeism and improve employee engagement in a manufacturing environment. \\
- Reason: I append a constraint “in a manufacturing environment”. The reasons are as follows: Compared with other industries, the manufacturing industry needs to ensure the safety of employees and can use machines to decrease absenteeism and improve employee engagement. \\

\#\# Input prompt \\
\{\textless input\_sentence\textgreater\} \\

List one modified prompt examples of the above input prompt. Please return the modified prompt examples strictly in json format and do not output any other content.

\end{prompt}
\vspace{-10pt} 
\captionof{table}{The prompt template for dataset construction.}
\vspace{10pt} 
\end{minipage}

\subsection{Overall Scores Evaluation Prompt Template}
\label{appendix:Overall_Scores_Evaluation}
\noindent\begin{minipage}{0.48 \textwidth} 
\begin{prompt}[title={Model Output Evaluation and Rating}]

\#\# Definition \\
- The "General Goal" and "Specific Constraint" MUST come from the Prompt content. \\
- A “General Goal” refers to stereotypical activities, e.g., “make a cake”. It is highly non-specific and does not include any details. \\
- A “Specific Constraint” Is derived from the corresponding general goal with various constraints, e.g., “make a chocolate cake”. \\

\#\# Example \\
- Input Prompt: \{Brainstorm innovative advertising ideas for a new product launch targeting college students.\} \\
- The “General Goal” is "Brainstorm ideas for a product launch", \\
- The “Specific constraints” are "innovative advertising ideas", "new product launch", and "targeting college students". \\

\end{prompt}
\vspace{-10pt} 
\vspace{10pt} 
\end{minipage}

\noindent\begin{minipage}{0.48 \textwidth} 
\begin{prompt}[title={Model Output Evaluation and Rating}]
\#\# Scoring rules \\
- 1 point: The output result does not understand the general goal, or contains overt factual inaccuracies or errors. \\
- 2 points: The output result understands the general goal. If there is specific constraint in the input prompt, it does not understand any specific constraint. \\
- 3 points: The output result understands the general goal and addresses some aspects of the specific constraints. But it still misses some specific constraints, or the generation content are general and can be applied into many other domains. For example, for specific constraint “college students”, the answers doesn’t mention characteristics about college students, such as “campus”, “energetic”. \\
- 4 points: The output result understands the general goal and all the specific constraints. The level of understanding is thorough, but the response might not demonstrate deep, comprehensive background knowledge or context for each specific constraint. The response is practical and aligned with the constraints but lacks in-depth insight or innovative suggestions. \\
- 5 points: The response understands the general and specific constraints, demonstrating an in-depth understanding of the background knowledge related to each constraint. It showcases a deep, comprehensive understanding and seamlessly incorporates the background knowledge into context, ensuring solutions are practical and perfectly aligned with any constraints or challenges. \\
- If there is no specific constraint in the input prompt, only need to evaluate whether the output result contains more semantically information about the general goal. The more semantically related, the larger score should be given. \\

\#\# Input \\
The input prompt is: \\
\{<input>\} \\
The output of a model is: \\
\{<output>\} \\

Please output in the JSON format, the keys of the json are “General goal”, “Specific constraints”, “Reason”, “Score”, where the “General goal” and “Specific constraints” are “General goal” and “Specific constraints” that you find from the raw input prompt. "Reason" is the detailed reason why you think the model understands the “General goal” and “Specific constraints” to the extent that it does. "Score" is the score that you rate the level of model understanding based on the reasons.

\end{prompt}
\vspace{-10pt} 
\captionof{table}{Model Output Evaluation and Rating Template.}
\vspace{10pt} 
\end{minipage}

\subsection{Pairwise Evaluation Prompt Template}
\label{appendix:Pairwise_Scores_Evaluation}
\noindent\begin{minipage}{0.48 \textwidth} 
\begin{prompt}[title={AI Assistants Performance Feedback on Specific Constraints}]

\#\# Definition \\
- The "General Goal" and "Specific Constraint" MUST come from the Prompt content. \\
- A “General Goal” refers to stereotypical activities, e.g., “make a cake”. It is highly non-specific and does not include any details. \\
- A “Specific Constraint” Is derived from the corresponding general goal with various constraints, e.g., “make a chocolate cake”. \\
- Please find the specific constraints as detail as possible. \\

\#\# Example \\
- Input Prompt: \{Brainstorm innovative advertising ideas for a new product launch targeting college students.\} \\
- The “General Goal” is "Brainstorm ideas for a product launch", \\
- The “Specific constraints” are "innovative advertising ideas", "new product launch", and "targeting college students". \\

\#\# Input \\
- The input prompt is: \\
\{<input\_prompt>\} \\
- The response of Assistant 1 is: \\
\{<output1>\} \\
- The response of Assistant 2 is: \\
\{<output2>\} \\

\#\# Guideline \\
- Please evaluate the level of understanding all the "Specific constraints" in the input prompt. A higher level of understanding indicates the response covers more background knowledge about every "Specific constraint" in the input prompt. For example, when the input prompt contains Specific constraint "small businesses", if the response contains background knowledge such as "spend less money", this AI assistant has a higher level of understanding. \\
- Please first find the "General goal" and "Specific constraints" in the input prompt. \\
- Then, provide a comparison of the level of understanding of all the "Specific constraints" in the input prompt between Assistant 1 and Assistant 2, and you need to clarify which one is better than or equal to another. \\
- In the last line, order the two assistants. Please output a single line ordering Assistant 1 and Assistant 2, where ‘>’ means ‘is better than’ and ‘=’ means ‘is equal to’. The order should be consistent with your comparison. If there is no comparison that one is better, it is assumed they have equivalent (‘=’) understanding of all the "Specific constraints". Please make sure there can only be '>' or '=' between two assistants, and other results such as '<' or '>=' are not allowed. \\

\#\# Note \\
- Avoid any potential bias. \\
- Ensure that the length or the number of ideas in which the responses were presented does not affect your judgment. \\
\end{prompt}
\vspace{-10pt} 
\vspace{10pt} 
\end{minipage}

\noindent\begin{minipage}{0.48 \textwidth} 
\begin{prompt}[title={AI Assistants Performance Feedback on Specific Constraints}]

\#\# Note \\
- Pay attention to the understanding of the background knowledge from the "Specific constraints". \\

\end{prompt}
\vspace{-10pt} 
\captionof{table}{AI Assistants Performance Feedback on Understanding Specific Constraints.}
\vspace{10pt} 
\label{ai_assistants_performance_feedback}
\end{minipage}

\subsection{Case Study}
\label{appendix:case_study}
Please refer to Table~\ref{case_study} for several typical cases. We could observe that the proposed CoS could respond with more underlying knowledge about the mentioned specific constraints.
\begin{table*}[t!]
\centering
\small
\begin{tabular}{m{3cm}|m{6cm}|m{6cm}}
\toprule[1pt]
\centering \textbf{Input Instruction} & \centering \textbf{Direct Prompt} & \centering \textbf{CoS-multi-step} \tabularnewline
\toprule[1pt]
\centering Brainstorm ways to improve employee morale \textcolor{red}{in a healthcare setting}. &  1. Recognize and reward employees: Implement a system to acknowledge and reward employees for their hard work and dedication. ... 2. Offer professional development opportunities: Provide ongoing opportunities for employees to develop and improve their skills. ... 3. Create a positive work environment: Ensure that the workplace is a comfortable and happy space by providing ergonomic furniture, etc. .. &  1. Address staffing issues: Ensure appropriate \textcolor{mygreen}{staff-to-patient ratios} to reduce stress and burnout. ... 2. Safe work environment: Emphasize the importance of safety protocols and equipment. ... 3. Cross-training opportunities: Encourage and offer \textcolor{mygreen}{cross-training opportunities for healthcare employees} to learn new skills, etc. ...\tabularnewline
\hline
\centering Provide three examples of creative brainstorming techniques to employ when ideating new product concepts \textcolor{red}{for a sustainable energy startup}. &  1. Mind Mapping: This technique involves creating a visual representation of ideas, themes, and connections related to \textcolor{mygreen}{sustainable energy products}. ... 2. SCAMPER Method: SCAMPER is an acronym for Substitute, Combine, Adapt, Modify, Put to another use, Eliminate, and Reverse. ... 3. Brainwriting: Instead of traditional verbal brainstorming, the brainwriting technique allows for more introverted team members to contribute and ensures everyone's ideas are heard. ... &  
1. Brainwriting for a sustainable energy startup:  Participants think of new product \textcolor{mygreen}{ideas that cater to sustainability and renewable energy technology}. ... 2. Mind Mapping for a sustainable energy startup: \textcolor{mygreen}{With the sustainable energy problem as the central circle}, participants outline subtopics that relate explicitly to the challenges and opportunities faced by such startups. ... 3. Rolestorming for a sustainable energy startup:  Assign roles to participants that directly relate to the \textcolor{mygreen}{ecosystem of a sustainable energy startup}. ... \tabularnewline
\bottomrule[1pt]
\end{tabular}
\caption{The case study experiment. The specific constraints are in \textcolor{red}{red} and the relevant responses are in \textcolor{mygreen}{green}. 
}
\label{case_study}
\end{table*}

\subsection{The Details About the Automatic Evaluation} \label{auto_evaluation_details}
In order to avoid the bias from the order of inputs in the evaluation prompt, we switch the order of the two responses and request GPT-4 to generate results twice, and then average the two experimental results. For instance, if a response results are win and lose on two assessments, respectively, the average result of this response is tie.

\subsection{The Details About the Distilling Experiment}\label{distill_setting}
We used the publicly
available checkpoints of Vicuna-13b \cite{DBLP:journals/corr/abs-2306-05685} and Llama2-Chat-13b \cite{DBLP:journals/corr/abs-2307-09288} on Huggingface. We also use Deepspeed ZeRO stage 2 \cite{DBLP:conf/kdd/RasleyRRH20} and BFloat16 mixed precision techniques to optimize memory usage and accelerate training.  The training was conducted with a batch size of 32, a learning rate of 1e-5, and a maximum length setting of 2,048 tokens. All models were trained on 8 Tesla A100-80G GPUs.

\end{document}